\pdfoutput=1

\documentclass[11pt]{article}

\usepackage{acl}

\usepackage{times}
\usepackage{latexsym}

\usepackage[T1]{fontenc}

\usepackage[utf8]{inputenc}

\usepackage{microtype}

\usepackage{float}
\usepackage{subcaption}
\usepackage{graphicx}
\usepackage{xcolor}
\usepackage{graphics}
\usepackage{tabularx}
\usepackage{multirow}
\usepackage{booktabs}
\usepackage{amsmath}
\usepackage{bbold}
\usepackage{bm}
\usepackage{relsize}
\usepackage[para]{threeparttable}

\title{Self-Generated In-Context Learning: Leveraging Auto-regressive Language Models as a Demonstration Generator}

\author{Hyuhng Joon Kim$^\dagger$, Hyunsoo Cho$^\dagger$, Junyeob Kim$^\dagger$,  \\
 \textbf{Taeuk Kim$^{\ddagger}$, Kang Min Yoo$^{\dagger\mathsection\mathparagraph}$, Sang-goo Lee$^\dagger$}\\
 $^\dagger$Seoul National University,
 $^\ddagger$ Hanyang University, $^\mathsection$NAVER AI Lab, $^\mathparagraph$NAVER CLOVA \\
 \texttt{\{heyjoonkim, johyunsoo, juny116, sglee\}@europa.snu.ac.kr}\\
 \texttt{kimtaeuk@hanyang.ac.kr} \\ 
 \texttt{kangmin.yoo@navercorp.com}\\
 }

\begin{document}
\maketitle
\begin{abstract}
Large-scale pre-trained language models (PLMs) are well-known for being capable of solving a task simply by conditioning a few input-label pairs dubbed demonstrations on a prompt without being explicitly tuned for the desired downstream task.
Such a process (i.e., in-context learning), however, naturally leads to high reliance on the demonstrations which are usually selected from external datasets.
In this paper, we propose self-generated in-context learning (\textbf{SG-ICL}), which generates demonstrations for in-context learning from PLM itself to minimize the reliance on the external demonstration.
We conduct experiments on four different text classification tasks and show SG-ICL significantly outperforms zero-shot learning and is generally worth approximately 0.6 gold training samples. 
Moreover, our generated demonstrations show more consistent performance with low variance compared to randomly selected demonstrations from the training dataset.

\end{abstract}

\section{Introduction}

The scale of pre-trained language models (PLMs) is ever-growing as they tend to deliver more meaningful results with larger models and have reached the scale of hundreds of billions.
However, transferring such large-scale PLMs with the traditional method i.e., fine-tuning, is problematic as it entails an immense cost to train and store parameters for an individual task.
Numerous branches of work have been proposed to circumvent such issues, such as Adapters \cite{Houlsby2019ParameterEfficientTL}, LoRA \cite{hu2021lora}, and in-context learning (ICL) \cite{brown2020language}.

Among others, ICL is in the limelight as it derives answers only from the internal knowledge of PLMs without any parameter updates.
Specifically, ICL \textit{learns} to solve a task simply by conditioning a few input-label pairs dubbed \textbf{demonstrations} on a prompt, which serves to give contexts regarding the downstream task during the inference phase, allowing PLMs to solve the tasks better.
The working principle of ICL intuitively leads to high reliance on the demonstrations, and performance deeply varies depending on the assortment of the demonstrations.

Many lines of work tackled the issue of ICL's high reliance on the demonstration. For instance, \citet{lu2021fantastically} shown in-context learning suffers from the order sensitivity of the demonstrations. \citet{zhao2021calibrate} introduces a contextual calibration procedure to reduce the variance across different choices of demonstrations.
\citet{rubin2021learning} suggests demonstration selection by retrieving in-context samples.
Notably, \citet{liu-etal-2022-makes} showed that selecting a demonstration that has a high correlation with the test input can improve performance.

Motivated by previous research considering the limits of ICL's working process, we tried to solve the following research question: \\
\indent
\textit{1. Can we eliminate the dependency on the training dataset by generating demonstrations?} \\
\indent
\textit{2. If so, how can we create demonstrations with high input-demonstration correlation?}
 
\begin{figure*}[ht]
    \centering
    \includegraphics[width=0.99\textwidth]{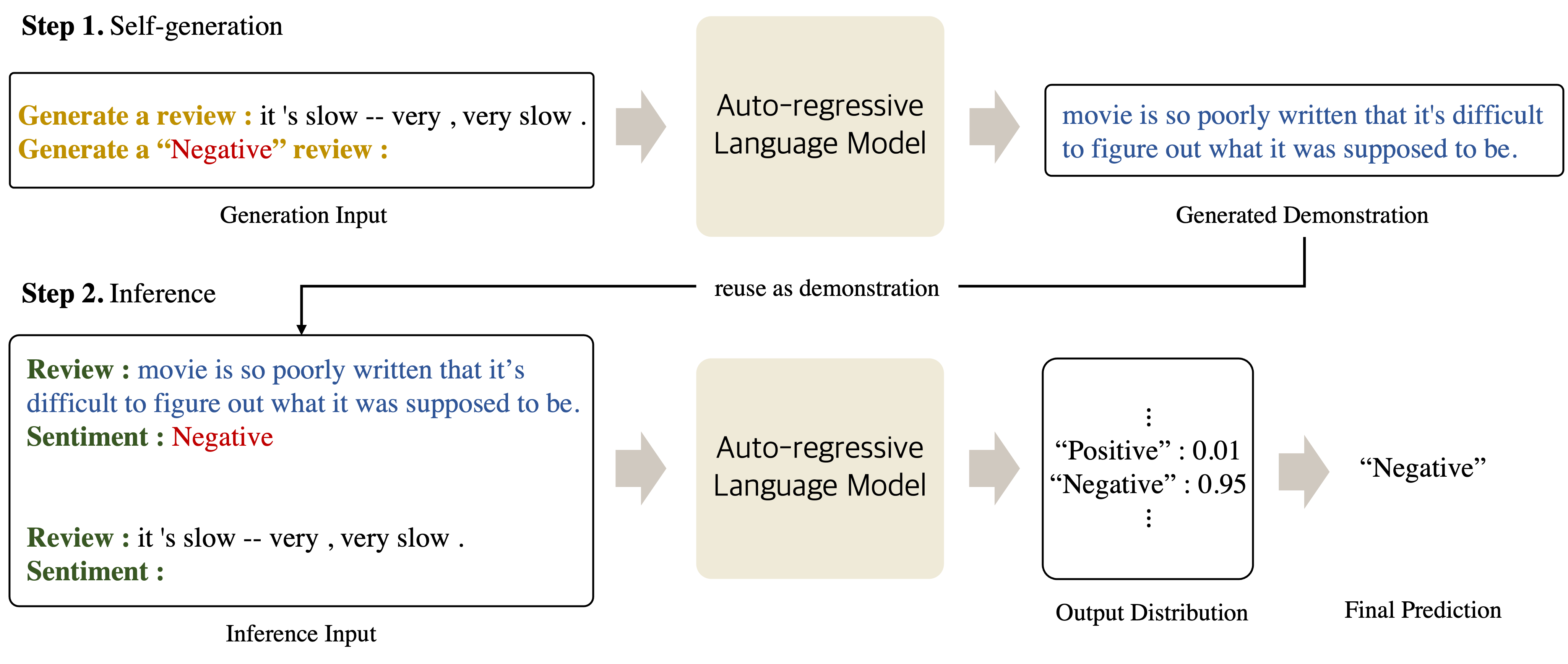}
    \caption{Overall process of SG-ICL. 
    Texts in yellow are manually designed prompts for generation and the texts in red are  expected class for generated demonstration. 
    Demonstrations (colored in blue) are generated in the self-generation step and are reused for in-context learning in the inference step. Texts in green are manually designed inference prompts.}
    \label{fig:overall_process}
\end{figure*}

To this end, we propose a novel method termed \textit{self-generated in-context learning} (\textbf{SG-ICL}) which generates demonstrations by leveraging the superiority of PLMs generative abilities \cite{adiwardana2020towards, brown2020language, shwartz-etal-2020-unsupervised, ye2022zerogen}.
To the best of our knowledge, this is the first study to utilize PLMs to create demonstrations for ICL.
SG-ICL consists of two operation steps: the self-generation step and the inference step.
In the self-generation step, we generate demonstrations for each class in the downstream task by conditioning on the current test input and class information with a simple manually designed template.
By giving conditions about the current input, PLM can generate demonstrations with a high input-demonstration correlation which is more befitted for ICL.
Then, the inference step performs ICL with generated demonstrations from the previous step which eliminates the requirement for training data or manual selection from training data.

We evaluate our method in four different natural language understanding (NLU) tasks, including sentiment classification and natural language inference. 
Through extensive experiments, we show that SG-ICL significantly outperforms zero-shot learning methods and is generally worth approximately 0.6 gold training samples.
Moreover, our generated demonstrations show more stable performance with low variance compared to randomly selected demonstrations from training dataset.

\section{Method }
\subsection{Few-shot Learning}
Given a PLM $ P $, our objective is to solve a classification task $ D^{test} = (X^{test}, Y^{test}) $. 
A natural language template $ T(\cdot) $ is provided, containing additional information about the downstream task. 
A limited number of training data $ D^{train} = (X^{train}, Y^{train}) $ is available as demonstration.
Inference input is generated by concatenating $ k $ training samples and the test input with the template.
Additionally, we define a verbalzier $ V( \cdot ) $ \cite{schick2020exploiting} which maps each class $ y_i \in Y $ to a pre-defined token. 
The final prediction is made by selecting the class with the highest probability for the mapped token :
\begin{equation}
    \begin{aligned}
        p(y_i \; | \; x^{test}_i) = P(V(y_i)) \; | \; T(x^{train}_1, y^{train}_1),\\
        ...,\; T(x^{train}_k, y^{train}_k),\; T(x^{test}_i)) 
    \end{aligned}
\end{equation}

Zero-shot learning is a special case of few-shot learning where the number of training data $ k = 0 $.

\begin{figure*}[ht]
    \centering
    \includegraphics[width=0.25\textwidth]{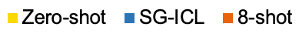} \\
    \begin{subfigure}[b]{0.48\textwidth}
        \centering
        \includegraphics[width=0.99\textwidth]{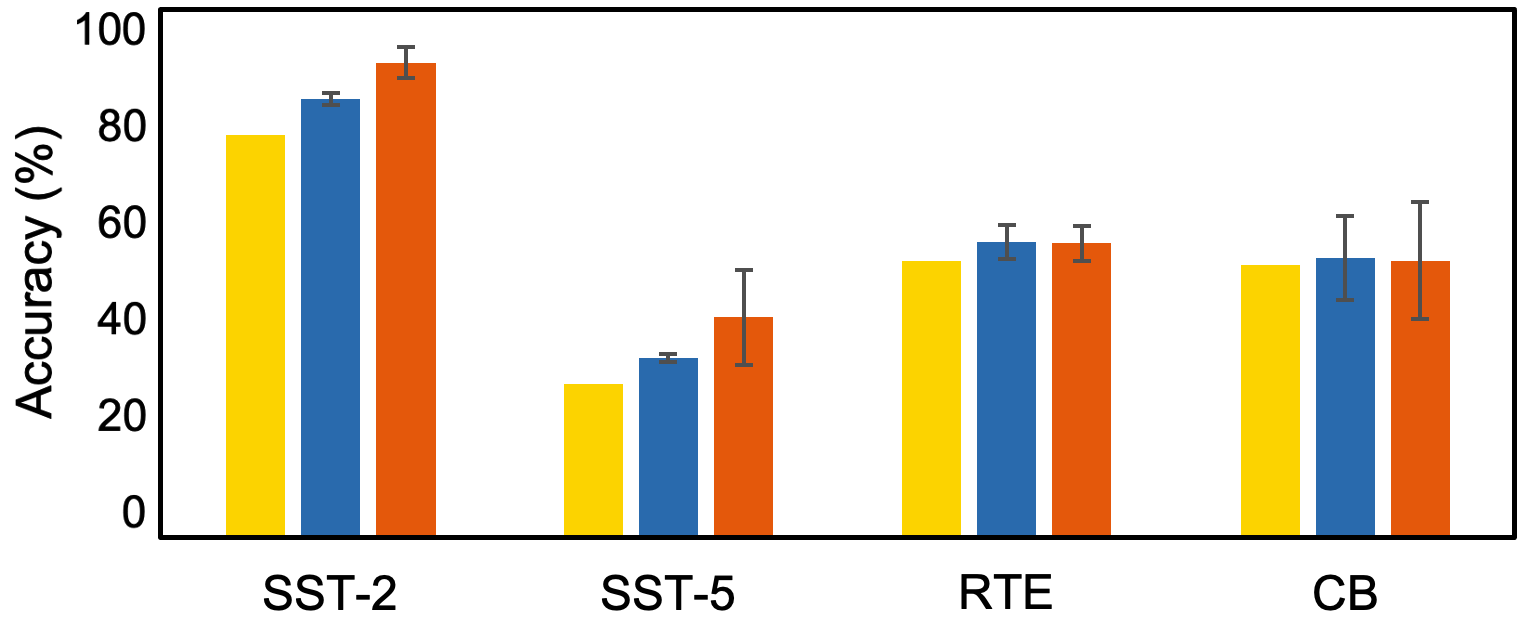}
        \caption{Results with manual inference template.}
        \label{fig:main_results_manual}
    \end{subfigure}
    \begin{subfigure}[b]{0.48\textwidth}
        \centering
        \includegraphics[width=0.99\textwidth]{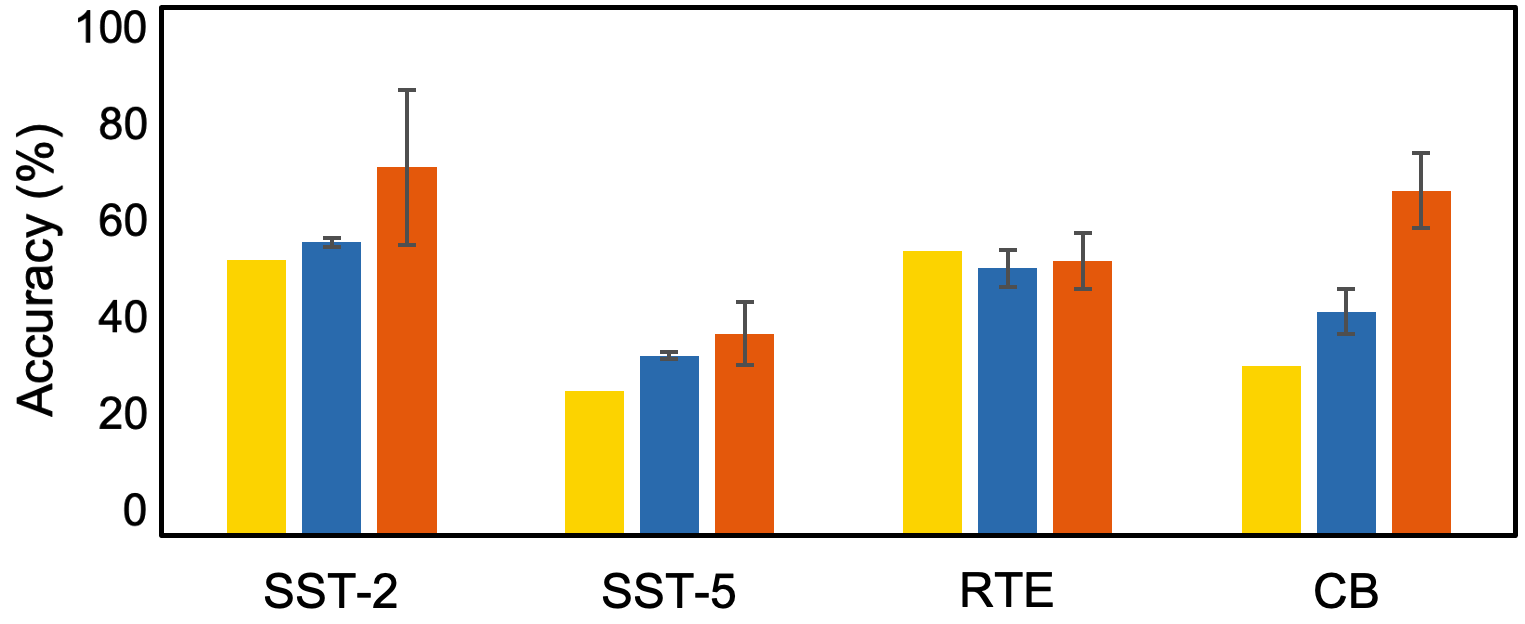}
        \caption{Results with minimal inference template.}
        \label{fig:main_results_minimal}
    \end{subfigure}
    \caption{Main results of our experiments. Figure \ref{fig:main_results_manual} and Figure \ref{fig:main_results_minimal} compares SG-ICL with zero/few-shot learning. Two settings are exactly the same except the inference template. SG-ICL outperforms zero-shot learning consistently. Notice that SG-ICL has significantly low variance compared to few-shot learning. }
    \label{fig:main_results}
\end{figure*}

\subsection{Self-generated In-context Learning}
SG-ICL can be divided into two steps : the self-generation step and the inference step. In the first step, we generate demonstrations conditioned on the test input and a specific class. This way we can generate demonstrations highly correlated with the test input. 
Details about the generation methods will be further discussed in Section \ref{sec:abl-conditioning}.
In the second step, we use the self-generated samples as a demonstration for in-context learning. The overall process of SG-ICL is visualized in Figure \ref{fig:overall_process}.

\medskip
\noindent
\textbf{Self-generation Step } In the self-generation step, we generate in-context sample $ s_i $.
Specifically, generation template $ G(\cdot) $ is defined, which takes the test instance $ x^{test}_i \in X^{test} $ and a class token $ V(y_i) $ as an input. 
The PLM $ P $ takes the generation input  $ G(x^{test}_i, V(y_i)) $ and generates in-context sample $ s_i $.

For single-sentence tasks (e.g., SST-2 and SST-5), the generation input can be defined as $ G(x^{test}_i, V(y_i))$. The generated in-context sample $ x^{gen}_i $ would be pairs of $ (s_i, y_i) $. 
In sentence-pair tasks (e.g., RTE and CB) consisting of sentence-pair inputs $ x_{i,1}, x_{i,2} $, the generation input can be defined as $ G(x_{i,1}, x_{i,2}, V(y_i)) $. In this case, generated in-context sample $ x^{gen}_i $ would be a set of $ (x_{i,1}, s_i, y_i) $.

\medskip
\noindent
\textbf{Inference Step } In the inference step, we use the generated samples as the demonstration for in-context learning. In detail, we take each generated samples $ x^{gen}_i $ and convert them using the inference template $ T(\cdot) $. The inference input is generated by concatenating all $ k $ generated samples and the test instance. The prediction is made in the same way as few-shot learning :
\begin{equation}
    \begin{aligned}
        p(y_i \; | \; x^{test}_i) = P(V(y_i)) \; | \; T(x^{gen}_1, y^{gen}_1),\\
        ...,\; T(x^{gen}_k, y^{gen}_k),\; T(x^{test}_i)) 
    \end{aligned}
\end{equation}

\section{Experiments }

\subsection{Experimental Setup}
\noindent
\textbf{Datasets and Metrics } We report results on four text classification datasets : sentiment classification with SST-2 \cite{socher2013recursive} and SST-5 \cite{socher2013recursive}, natural language inference with CB \cite{de2019commitmentbank} and RTE \cite{dagan2005pascal}.
We report accuracy for all tasks. All reported results are averaged over 5 different random seeds. 
See Table \ref{tab:datasets} in Appendix \ref{sec:appendix-dataset} for details about the datasets.

\medskip
\noindent
\textbf{Baselines } We compare SG-ICL with zero-shot learning and few-shot learning with 8 in-context samples. 2 different inference templates were used: minimal and manual. For inference templates and verbalizers, see Table \ref{table:inference_templates} in Appendix \ref{sec:appendix}.

\medskip
\noindent
\textbf{Models } The main experiments were done with GPT-J (6B) \cite{mesh-transformer-jax}, one of the largest publicly available auto-regressive models. We used the implementation and the pre-trained weights from Huggingface Transformers library \cite{wolf2019huggingface}.

\medskip
\noindent
\textbf{Generation Settings } For each test input $ x^{test}_i $, we self-generate 8 in-context samples. We use temperature sampling for generation \cite{hinton2015distilling} with temperature $ T=0.5$.
Details of generation templates are available in Table \ref{table:generation_templates} in Appendix \ref{sec:appendix}.

\subsection{Main results}
\label{sec:main_results}
Figure \ref{fig:main_results} is our main experimental results with the settings stipulated above.
We compare SG-ICL with zero/few-shot learning with the same number of in-context samples. 
We observed that SG-ICL performs significantly better than zero-shot learning, consistent across all four text classification tasks. 
This result is significant since both zero-shot learning and SG-ICL does not have any access to training data, but SG-ICL was able to gain improvements by using self-generated in-context samples.

Additionally, we can observe that the performance with SG-ICL is stable with very low variance. 
As in-context learning is highly dependent on the choice of the demonstration, its performance fluctuations are not negligible. SG-ICL alleviates this downside by generating an input conditioned sample highly correlated with the input instance and provides stable performance. 

\subsection{Why condition on the input? }
\label{sec:abl-conditioning}
In this section, we show the effect of conditioning on the input instance during the generation process. 
To do so, we conduct a simple experiment comparing two types of generation methods: (1) conditioning only on the class and (2) conditioning on both the class and the input instances.
Previous work \cite{liu-etal-2022-makes} has shown that in-context samples semantically-similar to the input instance are more likely to serve as a better in-context sample, improving performance. 
Based on previous research, we conduct an experiment to see whether conditioning on the input instance has a significant impact on the performance gain. We first calculate the correlation between the generated sample and the input instance. We use the sentence embedding from \citet{reimers-2019-sentence-bert} and calculate the cosine similarity between the two instance. 
Table \ref{tab:similarity} shows the results on SST-2 and SST-5.
We can observe that samples generated conditioned on the input instance shows a higher correlation with the input instance.

\begin{table}[t]
    \centering
    \begin{tabular}{ccc}
    \toprule
    Method & SST-2 & SST-5 \\ \hline
    \begin{tabular}[c]{@{}c@{}}class conditioned \end{tabular} & 0.0689 & 0.0735  \\
    \begin{tabular}[c]{@{}c@{}}input-class conditioned\end{tabular} & \textbf{0.3051} & \textbf{0.3098} \\ \bottomrule
    \end{tabular}
    \caption{Cosine similarity of the input instance and the generated demonstration. Inputs with similar sentence embedding have higher cosine similarity. Generating samples conditioned on both the input and the class shows higher cosine similarity.}
    \label{tab:similarity}
\end{table}

\begin{figure}[t]
    \centering
    \includegraphics[width=0.45\textwidth]{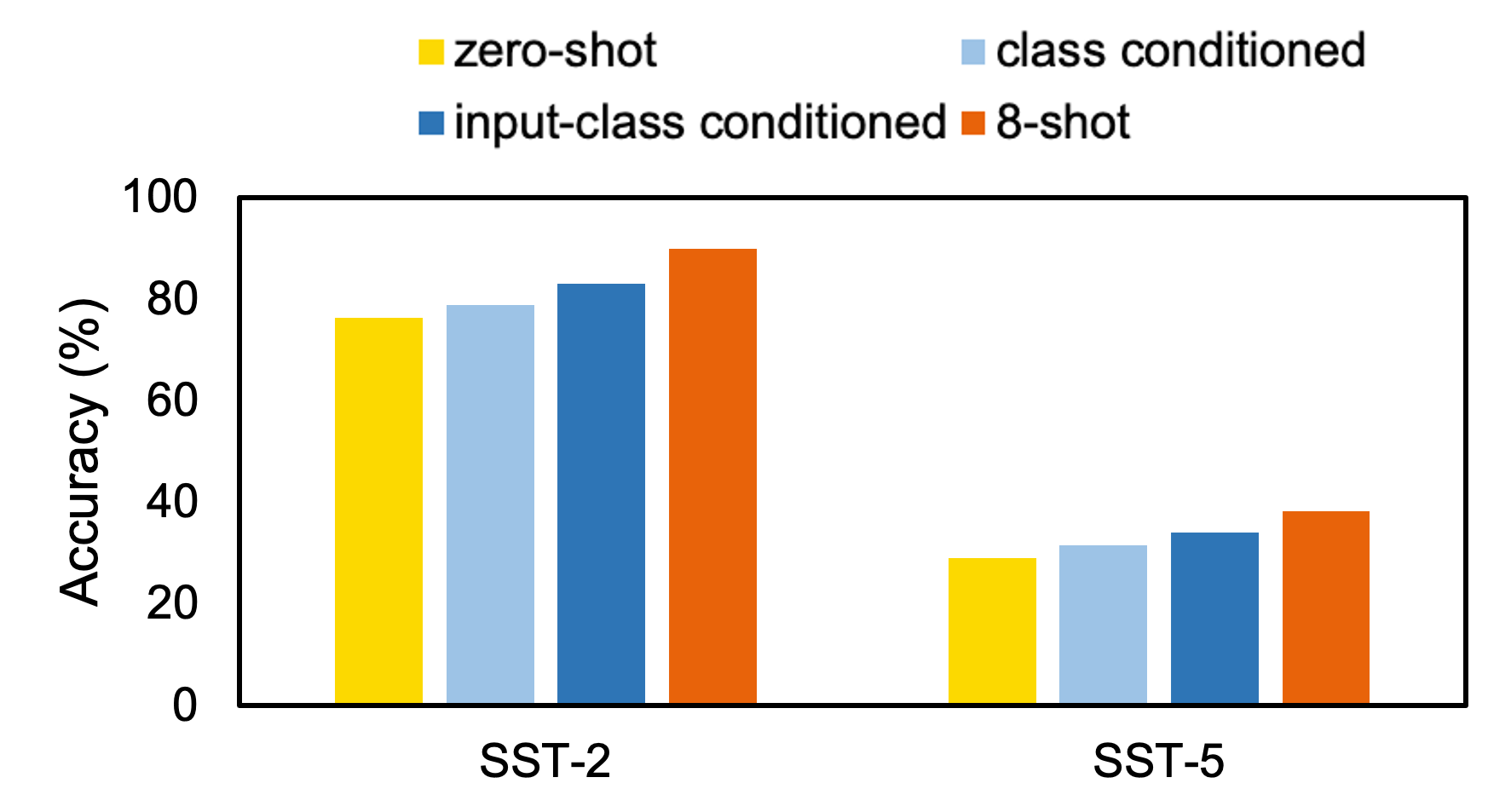}
    \caption{Results comparing two different generation methods alongside zero/few-shot in-context learning. Generating samples conditioned additionally on the input instance provides more performance gain.}
    \label{fig:ablation-3}
\end{figure}

To verify the correlation between the similarity and the downstream task performance, we report the downstream task performance of the two generation methods. The results are shown in Figure \ref{fig:ablation-3}. Regardless of the generation method, making use of the generated demonstrations provides performance gain. Additionally, we can observe in-context samples conditioned additionally on input instance performs better, aligning with the results in Table \ref{tab:similarity}.

\begin{figure}[h]
    \centering
    \begin{subfigure}[b]{0.45\textwidth}
        \includegraphics[width=0.99\textwidth]{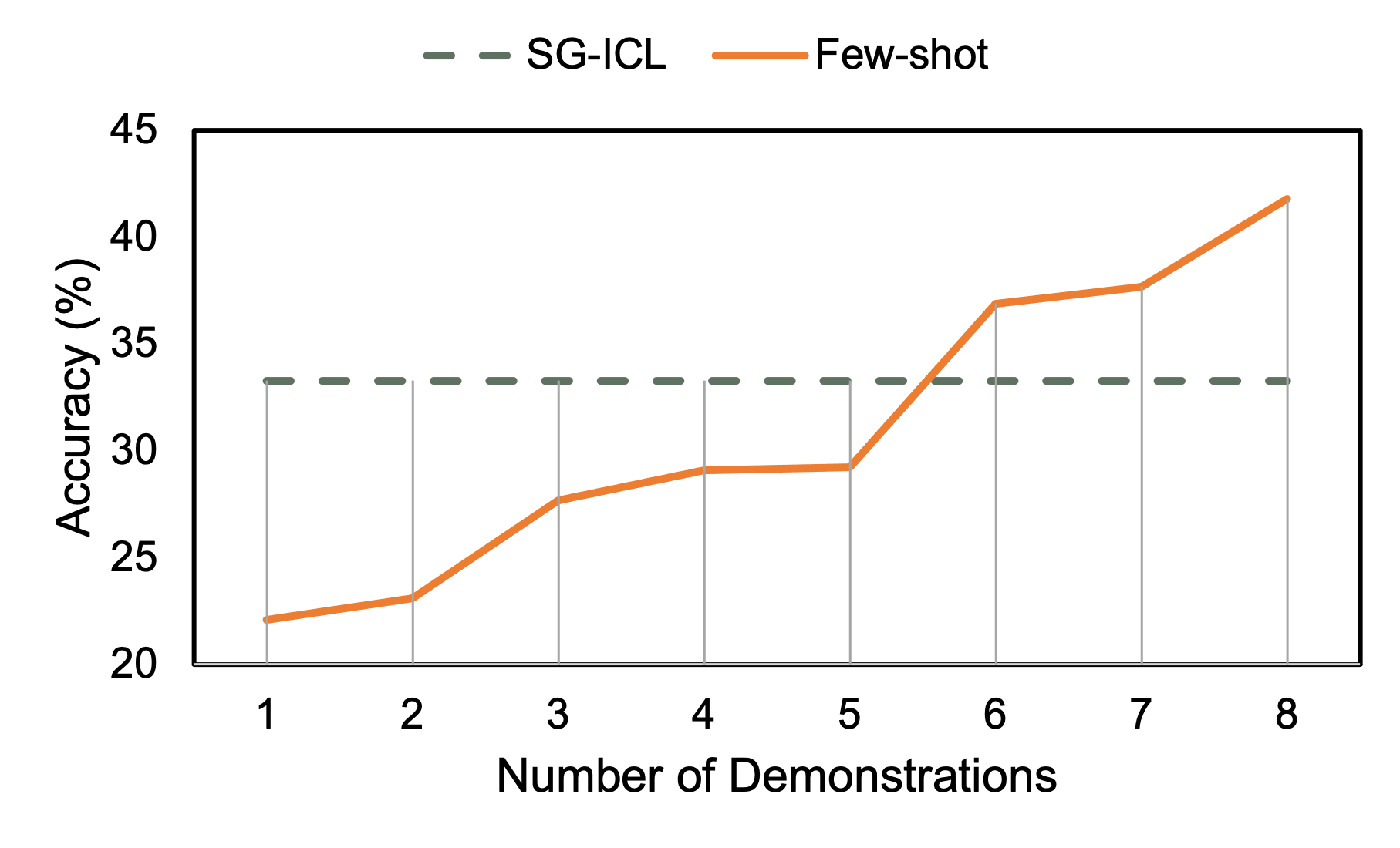}
        \caption{Results on SST-5.}
    \end{subfigure}
    \begin{subfigure}[b]{0.45\textwidth}
        \includegraphics[width=0.99\textwidth]{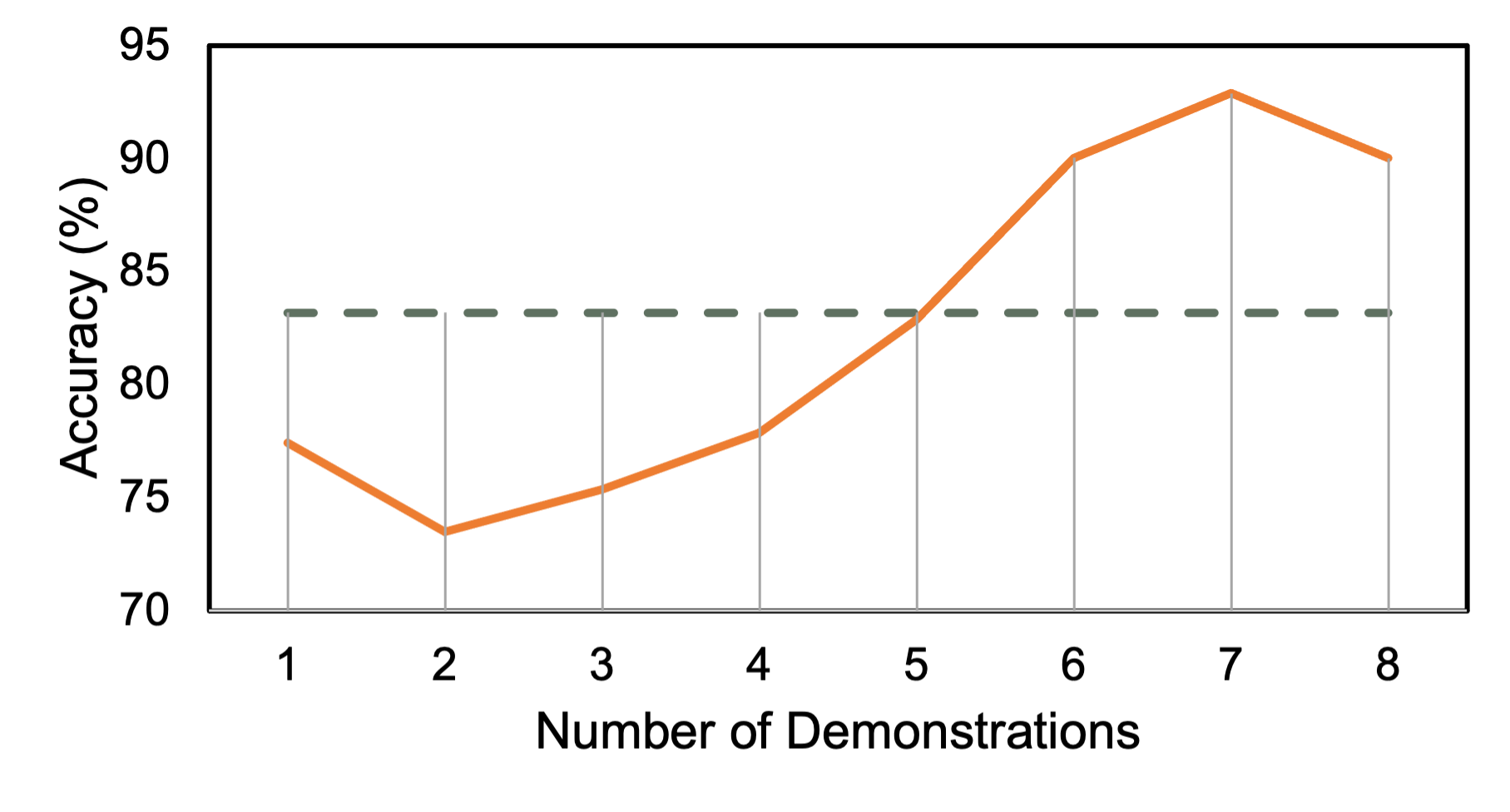}
        \caption{Results on SST-2.}
    \end{subfigure}
    \caption{Results comparing few-shot learning with various sample sizes with SG-ICL. SG-ICL outperforms few-shot learning with up to 5 in-context samples. }
    \label{fig:ablation-1}
\end{figure}

\subsection{How many in-context samples does self-generation worth?}
\label{sec:ablations}
We first analyze the significance of self-generated in-context samples in SG-ICL. To do so, we show few-shot learning performances with a varying number of in-context samples from 1 to 8 and compare it with SG-ICL. Figure \ref{fig:ablation-1} shows the results on two tasks: SST-2, SST-5. 
The results show that using 8 self-generated in-context samples can outperform few-shot learning with at most 5 gold training samples consistently. Experiments show one self-generated in-context sample worth about 0.6 gold training sample.

\section{Conclusion \& Future Work}
\label{sec:conclusion}
In this paper, we propose self-generated in-context learning (SG-ICL), generating in-context samples and reusing them as demonstrations.
We were able to generate quality demonstrations, eliminating the need for training data during in-context learning.
Our experiments on four text classification datasets show that SG-ICL can provide significant improvements in performance without the use of any training data. 
Moreover, SG-ICL show more stable performance with low variance compared to randomly selected demonstrations from training dataset.

As the quality of the generated samples is highly dependent on the generation abilities of the PLM, applying SG-ICL to larger PLMs would likely show significant improvements and is left as future work. Despite the positive results in natural language understanding tasks, applying SG-ICL to other tasks is not fully explored. Future work could include applying SG-ICL in various task domains. 

\section*{Acknowledgements}
This work was supported by SNU-Naver Hyperscale AI Center.

\bibliography{anthology,custom}

\appendix
\section{Dataset Details}
\label{sec:appendix-dataset}
Table \ref{tab:datasets} shows detailed statistics of the datasets used in the main experiment.

\begin{table}[h]
    \centering
    \resizebox{0.99\columnwidth}{!}{%
        \begin{tabular}{cccc}
        \toprule
            Dataset & \# of Train Set & \# of Validation Set & \# of Classes \\ \midrule
            SST-2 & 67,349 & 872 & 2 \\
            SST-5 & 8,544 & 2,210 & 5 \\
            RTE & 2,490 & 277 & 2 \\
            CB & 250 & 57 & 3 \\ \bottomrule
        \end{tabular}
    }
    \caption{Datasets used for experiments. }
    \label{tab:datasets}
\end{table}

\section{Templates for Generation and Inference}
\label{sec:appendix}
Table \ref{table:generation_templates} and Table \ref{table:inference_templates} shows details of the prompts and verbalizers used for inference and generation, respectively.

\begin{table*}[h]
\centering

    \begin{tabular}{cll}
    \toprule
    \textbf{Task} & \multicolumn{1}{c}{\textbf{Inference Template}} & \multicolumn{1}{c}{\textbf{Verbalizer}} \\ \midrule
    \multicolumn{1}{l}{Minimal} & \begin{tabular}[c]{@{}l@{}}a fast , funny , highly enjoyable movie .\\ \textcolor{blue}{positive}\end{tabular} & -  \\ \hline
    SST-2 & \begin{tabular}[c]{@{}l@{}}\textcolor{red}{Review : }a fast , funny , highly enjoyable movie .\\ \textcolor{red}{Sentiment : }\textcolor{blue}{positive}\end{tabular} & positive / negative \\ \hline
    SST-5 & \begin{tabular}[c]{@{}l@{}}\textcolor{red}{Review : }it 's worth taking the kids to .\\ \textcolor{red}{Sentiment :} \textcolor{blue}{great}\end{tabular} & \begin{tabular}[c]{@{}l@{}}terrible / bad/ \\ okay/ good / great\end{tabular} \\ \hline
    RTE & \begin{tabular}[c]{@{}l@{}}\textcolor{red}{Premise : }Dana Reeve, the widow of the actor Christopher Reeve, \\ has died of lung cancer at age 44, according to the Christopher \\ Reeve Foundation.\\ \textcolor{red}{Hypothesis : }Christopher Reeve had an accident.\\ \textcolor{red}{True or False?} \textcolor{blue}{false} \end{tabular} & true / false \\ \hline
    CB & \begin{tabular}[c]{@{}l@{}}\textcolor{red}{Premise :} It was a complex language. Not written down but \\ handed down. One might say it was peeled down.\\ \textcolor{red}{Hypothesis :} the language was peeled down\\ \textcolor{red}{Yes, No, or Neither?} \textcolor{blue}{yes} \end{tabular} & yes / no / neither \\ \bottomrule
    \end{tabular}
    
    \caption{Templates and verbalizers for inference. Texts in red are manually designed prompts and texts in blue are expected output for prediction. The rightmost column shows tokens mapped with each class. }

    \label{table:inference_templates}
\end{table*}

\begin{table*}[h]
\centering
    \begin{tabular}{cl}
    \toprule
    \textbf{Task} & \multicolumn{0}{c}{\textbf{Generation Template}} \\ 
    \midrule
    SST-2 & \begin{tabular}[c]{@{}l@{}}\textcolor{red}{Generate a review :} a fast , funny , highly enjoyable movie .\\ \textcolor{red}{Generate a }\textcolor{blue}{"negative" }\textcolor{red}{ review :}\end{tabular} \\ \midrule
    SST-5 & \begin{tabular}[c]{@{}l@{}}\textcolor{red}{Generate a review :} it 's worth taking the kids to .\\ \textcolor{red}{Generate a }\textcolor{blue}{"negative" }\textcolor{red}{ review :}\end{tabular} \\ \midrule
    RTE & \begin{tabular}[c]{@{}l@{}}\textcolor{red}{Premise : }Dana Reeve, the widow of the actor Christopher Reeve, has died of lung cancer \\ at age 44, according to the Christopher Reeve Foundation.\\ \textcolor{red}{Generate a Hypothesis :} Christopher Reeve had an accident.\\ \textcolor{red}{Generate a }\textcolor{blue}{"true" }\textcolor{red}{ Hypothesis :}\end{tabular} \\ \midrule
    CB & \begin{tabular}[c]{@{}l@{}}\textcolor{red}{Premise :} It was a complex language. Not written down but handed down. One might say \\it was peeled down.\\ \textcolor{red}{Generate a Hypothesis :} the language was peeled down\\ \textcolor{red}{Generate a }\textcolor{blue}{"neither" }\textcolor{red}{ Hypothesis :} \end{tabular} \\ 
    \bottomrule
    \end{tabular}
    
    \caption{Templates for self-generating in-context samples. Texts in red are manually designed prompts for generation and texts in blue are tokens representing the expected class.}
    
    \label{table:generation_templates}
\end{table*}

\end{document}